\documentclass[nohyperref]{article}

\usepackage{microtype}
\usepackage{graphicx}
\usepackage{subfigure} 
\usepackage{booktabs} 

\usepackage{hyperref}

\usepackage[accepted]{icml2022}

\usepackage{amsmath}
\usepackage{amssymb}
\usepackage{mathtools}
\usepackage{amsthm}

\usepackage{multirow}
\usepackage[figuresright]{rotating}
\usepackage{amsfonts}	
\usepackage{amsmath}
\usepackage{bm}
\usepackage{bbm}
\usepackage{xcolor}

\usepackage{bbold}
\usepackage{array}
\newcolumntype{x}[1]{>{\centering\arraybackslash}p{#1}}
\usepackage{tikz}
\usetikzlibrary{calc}
\usetikzlibrary{arrows,shapes,chains}
\usetikzlibrary{decorations.pathmorphing}

\usepackage[capitalize,noabbrev]{cleveref}

\theoremstyle{plain}
\newtheorem{theorem}{Theorem}[section]
\newtheorem{proposition}[theorem]{Proposition}

\theoremstyle{definition}

\theoremstyle{remark}

\usepackage[textsize=tiny]{todonotes}

\icmltitlerunning{Deep Squared Euclidean Approximation to the Levenshtein Distance}

\begin{document}

\twocolumn[
\icmltitle{Deep Squared Euclidean Approximation to the Levenshtein Distance \\
for DNA Storage}




\begin{icmlauthorlist}
\icmlauthor{Alan J.X. Guo}{cam}
\icmlauthor{Cong Liang}{cam}
\icmlauthor{Qing-Hu Hou}{math}

\end{icmlauthorlist}

\icmlaffiliation{cam}{Center for Applied Mathematics, Tianjin University, Tianjin, China;}
\icmlaffiliation{math}{School of Mathematics, Tianjin University, Tianjin, China}

\icmlcorrespondingauthor{Alan J.X. Guo}{jiaxiang.guo@tju.edu.cn}

\icmlkeywords{Machine Learning, ICML}

\vskip 0.3in
]



\printAffiliationsAndNotice{}  

\begin{abstract}
	Storing information in DNA molecules is of great interest 
	because of its advantages in longevity, high storage density, 
	and low maintenance cost. 
	A key step in the DNA storage pipeline
	is to efficiently cluster the retrieved DNA sequences 
	according to their similarities. 
	Levenshtein distance is the most suitable metric on the similarity between 
	two DNA sequences, 
	but it is inferior in terms of computational complexity 
	and less compatible with mature clustering algorithms.
	In this work, we propose a novel deep squared Euclidean embedding 
	for DNA sequences using
	Siamese neural network, squared Euclidean embedding, and 
	chi-squared regression. 
	The Levenshtein distance is approximated
	by the squared Euclidean distance between the embedding vectors, 
	which is fast calculated and clustering algorithm friendly. 
	The proposed approach is analyzed theoretically and experimentally.
	The results show that the proposed embedding is efficient and robust.
\end{abstract}

\section{Introduction}
\label{sec:introduction}
With the increasing demand for storing information, scientists have been focusing on finding advanced storage media.
DNA molecules, which carry most of the genetic information \emph{in vivo}, 
have been tried as a new storage medium
\cite{church2012next,goldman2013towards,grass2015robust,erlich2017dna,organick2018random,dong2020dna,chen2021artificial}. 
Studies have shown that the DNA storage has high storage density, low maintenance cost, and fast parallel replication \cite{ping2019carbon,dong2020dna}. 

A typical pipeline for storing information in synthetic DNA molecules includes the following steps \cite{dong2020dna}. 
The binary data are firstly encoded as strings on the alphabet $\{\mathrm{A,C,G,T}\}$.
Such strings are called \emph{reference}s and usually have a fixed length in range $100-200$ nucleotides, 
which is limited by the biochemical techniques used to manipulate long DNA sequences.
In the next step, the DNA molecules used as storage media are synthesized according to the references.
These molecules are amplified and carefully preserved to increase 
the chances of future retrieval of the original information. 
To retrieve the stored information, the DNA molecules are sequenced into a bucket of strings, 
which are called the \emph{read}s, on four bases $\{\mathrm{A,C,G,T}\}$.
Due to the aforementioned amplification procedure, 
the set of reads usually includes several copies of each reference at the same time. 
Also, the biochemical procedures on DNA molecules are not fully reliable, 
hence the reads can contain errors such as base insertions, deletions or substitutions \cite{blawat2016forward}.
Finally, 
the references are recovered from the reads and decoded to retrieve the original binary data. 

A straightforward way to recover references from the reads 
is to cluster the noisy reads by 
the Levenshtein distance \cite{levenshtein1966binary}, 
and select one reference from each cluster.
The Levenshtein distance between two strings, also known as the edit distance, 
is the minimum number of 
insertions, deletions, or substitutions required to modify one string to the other. 
Three issues arise when the size of stored information grows. 
First, 
the number of reads increases linearly with the amount of information stored, 
which leads to a squared increase in the computational complexity 
of a plain clustering algorithm \cite{rashtchian2017clustering}.
Second, 
the Levenshtein distance is not easy to calculate. 
It is shown that the Levenshtein distance cannot be computed in $O(n^{2-\epsilon}), \forall \epsilon > 0$, 
unless the strong exponential time hypothesis is false \cite{masek1980faster,backurs2015edit}. 
Third, 
most mature clustering algorithms are based on $\ell_p$ distance \cite{hartigan1979algorithm}; 
these algorithms may fail or need to be adapted for Levenshtein distance. 
These three issues render the mission of clustering a large number of reads by plain algorithms impossible within reasonable time. 
Researchers have tried to address the challenge of clustering a huge amount of reads. 
For example, 
to cluster billions of the reads efficiently, Rashtchian \emph{et al.} \cite{rashtchian2017clustering} 
proposed a distributed, agglomerative clustering algorithm, 
based on the fact that clusters of the reads from a DNA storage pipeline are well-separated in Levenshtein distance. 
In their work, tools like the $q$-gram distance \cite{ukkonen1992approximate} and locality sensitive hashing (LSH) \cite{har2012approximate}
were used for approximating the Levenshtein distance. 

In this paper, the embedding of DNA sequences is investigated 
to approximate the Levenshtein distance.
The basic idea is to map each DNA sequence to its embedding vector, 
and use the easy-calculated distance between the embedding vectors 
to approximate the Levenshtein distance between DNA sequences.
With the embedding of DNA sequences, 
the task of clustering the huge number of reads in the DNA storage pipeline 
can be simplified by the following two aspects.
First, the similarity between two DNA sequences can be computed more efficiently, 
because the computational complexity of the common distances, 
for example the $\ell_p$ distances, is $O(n)$, 
while the complexity of the Levenshtein distance is at least $O(n^{2-\epsilon})$.
Second, numerous efficient clustering algorithms can be introduced into the DNA storage pipeline 
without much effort, as most of the mature clustering algorithms take $\ell_p$ distances 
as preferred distance measure. 

Several existing works have attempted to map the Levenshtein distance to the easily calculated approximations \cite{hanada2017practical}. 
In \cite{charikar2006embedding,ostrovsky2007low,andoni2009overcoming}, 
the Levenshtein distance on permutations 
or $\{0,1\}$-words was embedded into $\ell_p$ with low distortion. 
In \cite{chakraborty2016streaming}, 
the CGK-embedding was proposed, which is a randomized injective embedding of the Levenshtein distance into 
the Hamming distance; the application of this algorithm includes \cite{zhang2017embedjoin}.
The $q$-gram-based methods record the frequency distribution of the $q$-grams in strings and use it to approximate the Levenshtein distance, 
which include \cite{ukkonen1992approximate, bar2004approximating, sokolov2007vector}. 
Without the Levenshtein distance, 
more works on string similarity join have been proposed \cite{wang2014hashing, yu2016string}; 
most of these works are based on LSH, to name a few, 
\cite{buhler2001efficient, datar2004locality, rasheed2012efficient,yuan2014hash}. 

Artificial neural networks \cite{lecun2015deep} were also utilized to 
provide sequence embeddings to compute approximations of the Levenshtein distance. 
Deep methods are often data-driven and have shown powerful feature extraction capabilities \cite{lecun2015deep}.
Therefore, it is assumed that deep learning-based approaches 
are more flexible and perform better in 
metric embedding on data with underlying structure or distribution. 
In \cite{zhang2020neural}, the gated recurrent unit (GRU) \cite{cho2014learning}, 
which is a famous variation of the recurrent neural network (RNN) \cite{rumelhart1985learning}, 
was considered as a embedding function. 
The GRU was trained with a three-phased procedure and the 
triplet loss \cite{schroff2015facenet} was engaged for similarity capturing. 
In \cite{dai2020convolutional}, the researchers proved one-hot embedding and max-pooling preserve a bound on Levenshtein 
distance, and applied convolutional neural network (CNN) \cite{krizhevsky2012imagenet} for a convolutional embedding (CNN-ED). 
The authors also engaged the triplet loss to train their model. 
In \cite{corso2021neural}, the authors reformulated existing approaches 
and explored related tasks, such as hierarchical clustering and multiple sequence alignment, with their NeuroSEED. 

Considering that the exact Levenshtein distance costs at least $O(n^{2-\epsilon})$ in complexity, 
the following statement is straightforward.
\begin{proposition}\label{claim}
	If have no restrictions on the sequences, 
	we can not find a deep learning-based sequence embedding 
	that gives the Levenshtein distance 
	by a conventional distance of complexity $O(n)$.
\end{proposition}

Although it is impossible to calculate the exact Levenshtein distance 
between arbitrary sequences by deep embedding, 
data-driven models can still be applied if the data show some ``good'' characteristics.
In this paper, we use deep embedding to calculate the Levenshtein distance on 
a unique real-world dataset formed by the sequences from DNA storage experiments.
The engaged dataset has specific underlying structures and distributions, 
from which we expect it to enable an efficient data-driven deep embedding.
Features of the dataset are summarized as follows:
\begin{itemize}
	\item The reads use fixed and finite alphabet $\{\mathrm{A,C,G,T}\}$ or $\{\mathrm{A,C,G,T,N}\}$, 
	where the $\mathrm{N}$ represents failed bases in sequencing. 
	Also, the length distribution of the reads is centered around the length of 
	the references. 
	\item The distributions of base insertions, deletions and substitutions are stable 
	between two reads of the same cluster.
	In the study of \cite{blawat2016forward}, 
	it was shown that the error rates range from $0.1\%$ to $1\%$.
	In practice, two reads of about $150$ nucleotides from the same cluster 
	have little chance of having a Levenshtein distance greater than $30$.
	\item The two reads from different clusters are completely unrelated to each other, 
	which is ensured by the separated references from the DNA storage pipeline. 
\end{itemize}

With regard to these features of the DNA sequences 
and the characteristics of deep embedding networks, 
we establish the deep squared Euclidean embedding (DSEE) method to approximate 
the Levenshtein distance by squared Euclidean distance on the dataset of DNA sequences.
The proposed DSEE includes the following three main techniques. 
\begin{itemize}
	\item Instead of the triplet loss used in the related works \cite{dai2020convolutional,zhang2020neural}, 
	the simple strategy of Siamese neural network \cite{bromley1993signature} is used to 
	optimize the parameters of the embedding network. 
	This simple setting helps us focus on the embedding of the 
	Levenshtein distance. 
	\item 
	Instead of the $\ell_p$ distances, the 
	squared Euclidean distance between the embedding vectors is used to approximate 
	the Levenshtein distance between the original sequences. 
	This enables us to mathematically interpret the Levenshtein 
	distance from a completely new viewpoint. 
	We establish the connection between the Levenshtein distance and the degree 
	of freedom of the embeddings for the first time. 
	\item Instead of regressions that uses mean squared error or mean absolute error as 
	the optimization target, we propose the chi-squared regression, 
	which uses a loss function that simulates the relative entropy from the chi-squared distribution. 
	This loss function is skewed around 
	the ground truth Levenshtein distance, and coincides with the theoretical 
	distribution of the approximate distance in a more reasonable way. 
\end{itemize}
With the squared Euclidean embedding and the chi-squared regression, 
our work reveals the critical part of a neural network-based embedding 
of Levenshtein distance for the first time. 
To the best of our knowledge, these techniques are introduced and 
analyzed both theoretically and experimentally for the first time
in related fields. 

The remainder of this paper is structured as follows. 
In \cref{sec:notationAndDataset}, 
we provide some preliminaries on the focused embedding task. 
In \cref{sec:method}, the proposed DSEE is introduced in detail. 
In \cref{sec:experiments}, the experiments and ablation study are conducted. 
Some concluding remarks are presented in \cref{sec:conclusion}.

\section{Notation, Dataset, and Metrics on Performance}\label{sec:notationAndDataset}
\subsection{Notations and Problem Statement}
Given two reads $\bm{s}, \bm{t}$ on the alphabet $\{\mathrm{A,C,G,T,N}\}$, 
where the $\mathrm{N}$ represents the bases that failed in sequencing, 
the Levenshtein distance $d_{L}(\bm{s}, \bm{t})$ is the minimum number of insertions, deletions and substitutions  
required to convert $\bm{s}$ to $\bm{t}$. 
By its definition, the Levenshtein distance is symmetric on $\bm{s}, \bm{t}$, 
\emph{i.e.} $d_{L}(\bm{s}, \bm{t}) = d_{L}(\bm{t}, \bm{s})$. 

Let $\bm{u}, \bm{v}$ be two vectors from the $n$-dimensional real space $\mathbb{R}^n$, 
the squared Euclidean distance is the square of the Euclidean distance ($\ell_2$ distance) 
between $\bm{u}$ and $\bm{v}$, which is 
\begin{equation}
	d_{\ell_2^2} = \sum_{i=1}^{n} (u_i-v_i)^2. 
\end{equation} 
It must be noticed that the triangle inequality does not hold 
for the squared Euclidean distance, hence the squared Euclidean distance does not form a metric space in mathematics. 
For convenience, we will still use the word ``distance'' for all the distance-like 
definitions, no matter they are real distances or not. 

For a deep model that accepts only vectors, matrices, or tensors as its input data, 
we use the one-hot encoding to convert the DNA sequences into matrices. 
This was proved to be efficient in similar tasks \cite{dai2020convolutional}. 
In the rest of this paper, the sequence $\bm{s}$ and its one-hot embedding $\mathrm{onehot}(\bm{s})$ will not be 
distinguished and the same notation $\bm{s}$ will be used. 


As stated in the \cref{sec:introduction}, 
the main task of this paper is to find a method to 
convert the reads into embedding vectors and to approximate the Levenshtein distance 
using the squared Euclidean distance between the embedding vectors. 
Mathematically, it can be interpreted as finding a function $f$ that maps the reads $\bm{s},\bm{t}$ 
to the embedding vectors $\bm{u}=f(\bm{s}),\bm{v}=f(\bm{t})$, such that the approximation error
$|d_{L}(\bm{s},\bm{t}) - d_{\ell_2^2}(\bm{u},\bm{v})|$ is small. 

\subsection{Dataset}
The public data\footnote{The data can be accessed 
through \url{https://github.com/TeamErlich/dna-fountain}, 
\url{https://www.ebi.ac.uk/ena/data/view/PRJEB19305}, and 
\url{https://www.ebi.ac.uk/ena/data/view/PRJEB19305}.}
from the DNA-Fountain \cite{erlich2017dna}, a well-known DNA storage study, 
is used for training and testing. 
It offers both the references and reads from their DNA storage pipeline. 
Each of the references has fixed length $152$ nucleotides, 
while the lengths of the reads are variable, 
for insertions or deletions of bases may occur in the biochemical procedure. 
In order to construct the training and testing sets, 
the DNA-Fountain data is divided into two parts according to a partition on the reference set. 
Each part of the data contains its selected references and the reads originate from those references.
It is worth noting that the partition of the data keeps the training and testing sets disjoint. 

Take the construction of the training set as an example. 
The training set is a collection of tuples $((\bm{s},\bm{t}),d)$, 
where the $(\bm{s},\bm{t})$ is a pair of DNA sequences and $d$ is the 
Levenshtein distance between them $d = d_{L}(\bm{s},\bm{t})$. 
Recall that the purpose of approximating Levenshtein distances is to cluster DNA sequences, 
the proposed embedding method should emphasize the separation of small and large distances. 
In view of this, the training set collects both homologous pairs and non-homologous pairs,
where the sequences of homologous pair are generated from the same reference and have a small Levenshtein distance, 
while the sequences of non-homologous pair are completely independent and have a large Levenshtein distance. 
The ratio of homologous pairs to non-homologous pairs is set to $1:1$. 
Since a cluster of reads usually includes the identical copies of its source reference, 
the reference-read pairs are engaged as the homologous pairs in practice. 
For example, given a reference $\bm{r}^i$ and its reads $\{\bm{s}^i_j\}_j$, 
the homologous pairs are $\{(\bm{r}^i,\bm{s}^i_j)\}_j$ and 
the training set uses $((\bm{r}^i,\bm{s}^i_j),d_{L}(\bm{r}^i,\bm{s}^i_j))$ as its homologous samples. 
It is worth noting that, for any non-generative embedding method, 
the embedding vectors are the same for different runs on the same string. 
Therefore, pairs of identical sequences with Levenshtein distance $0$ 
are trival and should be screened out from the training set.
As for the non-homologous pairs, 
a random selection of two reads $(\bm{s}^{i_1}_{j_1},\bm{s}^{i_2}_{j_2})$ from different references 
$\bm{r}^{i_1},\bm{r}^{i_2}\,(i_1\neq i_2)$ 
will satisfy the requirement, 
and the training set uses the $((\bm{s}^{i_1}_{j_1},\bm{s}^{i_2}_{j_2}),d_{L}(\bm{s}^{i_1}_{j_1},\bm{s}^{i_2}_{j_2}))$ 
as its non-homologous samples. 
Finally, the training set is composed of samples $(\bm{s},\bm{t},d_L(\bm{s},\bm{t}))$ 
that include the homologous pairs and non-homologous pairs described above. 

Considering the differences in the read lengths, 
all the sequences are padded with $0$ at the end to achieve a fixed length of $160$ nucleotides. 
In practice, homologous pairs with large distances occur at much lower rates than pairs with small distances, 
therefore, the samples are balanced by duplicating according to their Levenshtein distances. 

\subsection{Metrics on Performance}\label{subsec:metrics}
A common metric used in regression tasks is the approximation error ($\mathrm{AE}$).
Given two sequences $\bm{s},\bm{t}$, let $\bm{u},\bm{v}$ 
be the two embedding vectors of these two sequences, respectively. 
In this paper, the mean absolute error (MAE) is engaged as the approximation error, 
\begin{equation}
    \mathrm{AE} = \mathrm{MAE} = \frac{1}{\# \mathrm{Te}}\sum_{(\bm{s},\bm{t})\in\mathrm{Te}}|d_L(\bm{s},\bm{t})-\hat{d}(\bm{u},\bm{v})|,
\end{equation}
where the $\mathrm{Te}$ is for the testing set, and the $\hat{d}(u,v)$ is the 
approximate distance between embedding vectors $\bm{u},\bm{v}$. 

In the specific task of this paper, 
the AE on the non-homologous pairs is less important 
than the AE on the homologous pairs. 
It is natural that 
the approximate distances are desired 
to be as accurate as possible on the homologous pairs. 
However, 
as stated in \cref{claim}, a global accurate approximation is impossible. 
A careful dipiction on the Levenshtein distance between non-homologous sequences, 
which is the Levenshtein distance from random sequence to random sequence, 
is complex and meaningless. 
In view of this, a biased AE on the testing set 
is also considered to be an important metric on the performance of the method.
Let $\mathrm{Te}_{h}$ be the collection of homologous pairs from the testing set $\mathrm{Te}$, 
the mean absolute error on $\mathrm{Te}_{h}$ ($\mathrm{MAE}_h$) is engaged as the biased approximation error ($\mathrm{AE}_h$), 
\begin{equation}
	\mathrm{AE}_h=\mathrm{MAE}_h = \frac{1}{\# \mathrm{Te}_h}\sum_{(\bm{s},\bm{t})\in\mathrm{Te}_h}
	|d_L(\bm{s},\bm{t})-\hat{d}(\bm{u},\bm{v})|,
\end{equation}
where the $\mathrm{AE}_h$ only takes the homologous pairs into account.

One of the motivations of the proposed method is to use approximate distances
to quickly determine whether two read have the same reference as a source. 
In the definition of the metric $\mathrm{AE}_h$, the approximation error on non-homologous 
pairs is dropped, but the proposed model should still have the ability to 
separate non-homologous and homologous pairs. 
To evaluate this ability, the metrics of classification tasks can be used. 
Given a threshold $K$, the testing samples can be divided into two groups by the approximate distance, 
which are
\begin{equation}
    \hat{d}(\bm{u},\bm{v}) \geq K
\end{equation}
for the predicted non-homologous pairs and 
\begin{equation}
    \hat{d}(\bm{u},\bm{v}) < K
\end{equation}
for the predicted homologous pairs. 
Compare the predicted classification results with the ground truth partition of testing samples into homologous and non-homologous, 
the overall accuracy ($\mathrm{OA}$) from classification tasks is used to evaluate separation ability of the proposed methods. 

\section{Method}\label{sec:method}
A typical deep model is a composite function of elementary neural network layers, 
which includes CNN, fully connected layer, \emph{etc.}. 
It has been proved to be efficient in various tasks, 
including feature extraction and embedding \cite{lecun2015deep}.
In the proposed DSEE, the deep model is engaged as the embedding network to transform the DNA sequences 
into their embedding vectors. 
In detail, the embedding network is trained as a part of the Siamese neural network; 
the squared Euclidean distance is used as the embedding distance;
and the model is trained by the chi-squared regression. 

\subsection{Siamese Neural Network}\label{subsec:twinmodel}
Siamese neural network is firstly introduced in \cite{bromley1993signature} 
for signature verification. 
It takes two inputs by two identical networks which share parameters,
and gives the similarity on the inputs by comparing the outputs of the networks. 

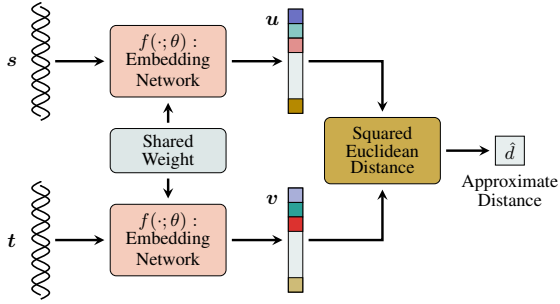
\begin{figure}
	\centering
{\linespread{1}
	\centering
	\tikzstyle{format}=[circle,draw,thin,fill=white]
	\tikzstyle{format_gray}=[circle,draw,thin,fill=gray]
	\tikzstyle{format_rect}=[rectangle,draw,thin,fill=white,align=center]
	\tikzstyle{arrowstyle} = [->,thick]
	\tikzstyle{network} = [rectangle, minimum width = 3cm, minimum height = 1cm, text centered, draw = black,align=center,rounded corners,fill=green_so,fill opacity=0.5,text opacity=1]
	\tikzstyle{training_batch} = [trapezium, trapezium left angle = 30, trapezium right angle = 150, minimum width = 3cm, text centered, draw = black, fill = cyan_so, fill opacity=0.3,text opacity=1,align=center]		
	\tikzstyle{class_features} = [trapezium, trapezium left angle = 30, trapezium right angle = 150, minimum width = 3cm, text centered, draw = black, fill = cyan_so, fill opacity=0.3,text opacity=1,align=center]
	\tikzstyle{pixel} = [rectangle, draw = black, fill = orange_so, fill opacity=0.5,text opacity=0,align=center]	
	\tikzstyle{pixel_red} = [rectangle, draw = black, fill = red_so, fill opacity=1,text opacity=0,align=center]	
	\tikzstyle{feature} = [rectangle, draw = black, fill = orange_so, fill opacity=0.3,text opacity=0,align=center,rounded corners]	
	\tikzstyle{feature_sfp} = [rectangle, draw = black, fill = violet_so, fill opacity=0.3,text opacity=0,align=center,rounded corners]					
	\tikzstyle{arrow1} = [thick, ->, >= stealth]
	\tikzstyle{arrow1_thick} = [thick, ->, >= stealth, line width=1.2pt]
	\tikzstyle{arrow2} = [thick, dashed, ->, >= stealth]
	\tikzstyle{thick_line} = [thick, line width=1.5pt]
	\tikzstyle{channel} = [fill=white,fill opacity = 0.7, rounded corners=3pt]
	\tikzstyle{channel_shadow} = [fill = gray_so, fill opacity = 0.1, rounded corners]
	\tikzstyle{channel_selected} = [fill = orange_so, fill opacity = 0.5]

	\tikzstyle{dna} = [decoration={coil}, decorate, thick, decoration={aspect=0, segment length=0.5*0.87cm, post length=0.,pre length=0.}]

	\begin{tikzpicture}[auto,>=latex', thin, start chain=going below, every join/.style={norm}]
		\definecolor{gray_so}{RGB}{88,110,117}
		\definecolor{lightgray_so}{RGB}{207,221,221}
		\definecolor{yellow_so}{RGB}{181,137,0}
		\definecolor{cyan_so}{RGB}{42,161,152}
		\definecolor{orange_so}{RGB}{203,75,22}
		\definecolor{green_so}{RGB}{133,153,0}
		\definecolor{red_so}{RGB}{220,50,47}
		\definecolor{magenta_so}{RGB}{211,54,130}
		\definecolor{violet_so}{RGB}{108,113,196}
		\definecolor{yellow_ad}{RGB}{242,228,201}
		\definecolor{pink_ad}{RGB}{242,182,160}
		\definecolor{green_ad}{RGB}{146,195,185}
		\definecolor{dgreen_ad}{RGB}{104,166,148}
		\definecolor{purple_ad}{RGB}{115,72,88}
		\definecolor{att_blue}{RGB}{185,233,248}
		\useasboundingbox  (0,0) rectangle (10*0.77,6.18*0.77);

		\scope[transform canvas={scale=0.77}]
		
		\coordinate (zero) at (0,0);
		\coordinate (upperhalf) at (0,4.635);
		\coordinate (half) at (0,3.09);
		\coordinate (lowerhalf) at (0,1.545);

		\coordinate (upperDnaCenter) at ($(upperhalf)+(0.8,0)$);
		\coordinate (lowerDnaCenter) at ($(lowerhalf)+(0.8,0)$);

		\draw[dna, decoration={amplitude=.15cm}] ($(upperDnaCenter)+(0,-0.95)$) -- ($(upperDnaCenter)+(0,1.02)$);
		\draw[dna, decoration={amplitude=-.15cm}] ($(upperDnaCenter)+(0,-1.02)$) -- ($(upperDnaCenter)+(0,0.95)$);
		\node at ($(upperDnaCenter)+(-0.5,0)$) {$\bm{s}$};

		\draw[dna, decoration={amplitude=.15cm}] ($(lowerDnaCenter)+(0,-0.95)$) -- ($(lowerDnaCenter)+(0,1.02)$);
		\draw[dna, decoration={amplitude=-.15cm}] ($(lowerDnaCenter)+(0,-1.02)$) -- ($(lowerDnaCenter)+(0,0.95)$);
		\node at ($(lowerDnaCenter)+(-0.5,0)$) {$\bm{t}$};

		\coordinate (upperModelCenter) at ($(upperDnaCenter)+(2.2,0)$);
		\coordinate (lowerModelCenter) at ($(lowerDnaCenter)+(2.2,0)$);
		\filldraw[channel, fill=pink_ad] ($(upperModelCenter)+(-1,-0.618)$) rectangle ($(upperModelCenter)+(1,0.618)$);
		\node at ($(upperModelCenter)+(0,0.33)$) {\small{$f(\cdot;\theta):$}};
		\node at ($(upperModelCenter)+(0,-0.0)$) {\small{Embedding}};
		\node at ($(upperModelCenter)+(0,-0.33)$) {\small{Network}};
		\filldraw[channel, fill=pink_ad] ($(lowerModelCenter)+(-1,-0.618)$) rectangle ($(lowerModelCenter)+(1,0.618)$);
		\node at ($(lowerModelCenter)+(0,0.33)$) {\small{$f(\cdot;\theta):$}};
		\node at ($(lowerModelCenter)+(0,-0.0)$) {\small{Embedding}};
		\node at ($(lowerModelCenter)+(0,-0.33)$) {\small{Network}};
		\draw[arrow1_thick] ($(upperDnaCenter)+(0.25,0)$) -- ($(upperModelCenter)+(-1.1,0)$);
		\draw[arrow1_thick] ($(lowerDnaCenter)+(0.25,0)$) -- ($(lowerModelCenter)+(-1.1,0)$);

		\coordinate (weightsCenter) at ($(upperModelCenter)!0.5!(lowerModelCenter)$);
		\filldraw[channel, fill=lightgray_so] ($(weightsCenter)+(-1,-0.382)$) rectangle ($(weightsCenter)+(1,0.382)$);
		\node at ($(weightsCenter)+(0,0.17)$) {\small{Shared}};
		\node at ($(weightsCenter)+(0,-0.17)$) {\small{Weight}};
		\draw[arrow1_thick] ($(weightsCenter)+(0,0.382+0.1)$) -- ($(upperModelCenter)+(0,-0.618-0.1)$);
		\draw[arrow1_thick] ($(weightsCenter)+(0,-0.382-0.1)$) -- ($(lowerModelCenter)+(0,0.618+0.1)$);

		\coordinate (upperFeatureCenter) at ($(upperModelCenter)+(2.2,0)$);
		\coordinate (lowerFeatureCenter) at ($(lowerModelCenter)+(2.2,0)$);
		\filldraw[channel_selected,fill=lightgray_so] ($(lowerFeatureCenter)+(-0.125,-0.9)$) rectangle ($(lowerFeatureCenter)+(0.125,0.9)$);
		\filldraw[channel_selected,fill=yellow_so,fill opacity = 0.5] ($(lowerFeatureCenter)+(0,0)+(-0.125,-0.9)$) rectangle ($(lowerFeatureCenter)+(0.25,0.25)+(-0.125,-0.9)$);
		\filldraw[channel_selected,fill=violet_so,fill opacity = 0.5] ($(lowerFeatureCenter)+(0,1.55)+(-0.125,-0.9)$) rectangle ($(lowerFeatureCenter)+(0.25,1.8)+(-0.125,-0.9)$);
		\filldraw[channel_selected,fill=cyan_so,fill opacity = 1] ($(lowerFeatureCenter)+(0,1.3)+(-0.125,-0.9)$) rectangle ($(lowerFeatureCenter)+(0.25,1.55)+(-0.125,-0.9)$);
		\filldraw[channel_selected,fill=red_so,fill opacity = 1] ($(lowerFeatureCenter)+(0,1.05)+(-0.125,-0.9)$) rectangle ($(lowerFeatureCenter)+(0.25,1.30)+(-0.125,-0.9)$);

		\filldraw[channel_selected,fill=lightgray_so] ($(upperFeatureCenter)+(-0.125,-0.9)$) rectangle ($(upperFeatureCenter)+(0.125,0.9)$);
		\filldraw[channel_selected,fill=yellow_so,fill opacity = 1] ($(upperFeatureCenter)+(0,0)+(-0.125,-0.9)$) rectangle ($(upperFeatureCenter)+(0.25,0.25)+(-0.125,-0.9)$);
		\filldraw[channel_selected,fill=violet_so,fill opacity = 1] ($(upperFeatureCenter)+(0,1.55)+(-0.125,-0.9)$) rectangle ($(upperFeatureCenter)+(0.25,1.8)+(-0.125,-0.9)$);
		\filldraw[channel_selected,fill=cyan_so,fill opacity = 0.5] ($(upperFeatureCenter)+(0,1.3)+(-0.125,-0.9)$) rectangle ($(upperFeatureCenter)+(0.25,1.55)+(-0.125,-0.9)$);
		\filldraw[channel_selected,fill=red_so,fill opacity = 0.5] ($(upperFeatureCenter)+(0,1.05)+(-0.125,-0.9)$) rectangle ($(upperFeatureCenter)+(0.25,1.30)+(-0.125,-0.9)$);
		\node at ($(upperFeatureCenter)+(-0.4,0.7)$) {$\bm{u}$};
		\node at ($(lowerFeatureCenter)+(-0.4,0.7)$) {$\bm{v}$};

		\draw[arrow1_thick] ($(upperModelCenter)+(1.1,0)$) -- ($(upperFeatureCenter)+(-0.125-0.1,0)$);
		\draw[arrow1_thick] ($(lowerModelCenter)+(1.1,0)$) -- ($(lowerFeatureCenter)+(-0.125-0.1,0)$);

		\coordinate (upperEuclideanCenter) at ($(upperFeatureCenter)+(1.5,0)$);
		\coordinate (lowerEuclideanCenter) at ($(lowerFeatureCenter)+(1.5,0)$);
		\coordinate (euclideanCenter) at ($(upperEuclideanCenter)!0.5!(lowerEuclideanCenter)$);

		\filldraw[channel, fill=yellow_so] ($(euclideanCenter)+(-1,-0.573)$) rectangle ($(euclideanCenter)+(1,0.573)$);
		\node at ($(euclideanCenter)+(0,0.28)$) {\small{Squared}};
		\node at ($(euclideanCenter)+(0,0)$) {\small{Euclidean}};
		\node at ($(euclideanCenter)+(0,-0.28)$) {\small{Distance}};

		\draw[arrow1_thick] ($(upperFeatureCenter)+(0.125+0.1,0)$) -- ($(upperEuclideanCenter)+(0,0)$) -- ($(euclideanCenter)+(0,0.573+0.1)$);
		\draw[arrow1_thick] ($(lowerFeatureCenter)+(0.125+0.1,0)$) -- ($(lowerEuclideanCenter)+(0,0)$) -- ($(euclideanCenter)+(0,-0.573-0.1)$);

		\coordinate (outputCenter) at ($(euclideanCenter)+(2.2,0)$);
		\draw[arrow1_thick] ($(euclideanCenter)+(1.1,0)$) -- ($(outputCenter)+(-0.25-0.1,0)$);
		\filldraw[channel_selected,fill=lightgray_so] ($(outputCenter)+(-0.25,-0.25)$) rectangle ($(outputCenter)+(0.25,0.25)$);
		\node at ($(outputCenter)$) {\small{$\hat{d}$}};
		\node at ($(outputCenter)+(0,-0.25-0.3)$) {\small{Approximate}};
		\node at ($(outputCenter)+(0,-0.25-0.6)$) {\small{Distance}};
		\endscope
	\end{tikzpicture}	
}
\caption{\label{fig:siamese} Siamese neural network. 
The two DNA sequences $\bm{s},\bm{t}$ are mapped to respective embedding vectors
$\bm{u}, \bm{v}$
via two branches which share the same parameters. 
The approximate distance is calculated by the squared Euclidean distance between 
$\bm{u}$ and $\bm{v}$. }
\end{figure}

In DSEE, a Siamese structure is used for training the embedding model. 
Let $(\bm{s},\bm{t},d)$ be a sample from the training set, 
and $f(\cdot;\theta)$ be the embedding network, 
the Siamese neural network transforms the inputs by the procedure shown in \cref{fig:siamese}. 
The two DNA sequences $\bm{s},\bm{t}$ are firstly mapped to their respective embedding vectors
${\bm{u}} = f(\bm{s};\theta), {\bm{v}} = f(\bm{t};\theta)$
via two branches of the Siamese structure. 
Each branch of the Siamese network is fulfilled by a complete embedding network $f(\cdot;\theta)$;
and the two branches of the embedding networks share the same weight $\theta$. 
Between the two embedding vectors ${\bm{u}},{\bm{v}}$, 
the approximate distance is calculated by 
$\hat{d} = d_{\ell_2^2}({\bm{u}},{\bm{v}})$, where $d_{\ell_2^2}(\cdot,\cdot)$ is 
the function of squared Euclidean distance, 
and the $\hat{d}$ is the approximation of the Levenshtein distance 
between the DNA sequences $\bm{s},\bm{t}$. 

To learn a optimized $\hat{\theta}$, which enables the embedding network $f(\cdot;\hat{\theta})$, 
a cost or loss function $\mathcal{L}(\bm{d},\hat{\bm{d}};\theta)$ is defined by the ground truth Levenshtein distances $\bm{d}$ 
and the approxiamte distances $\bm{\hat{d}}$ on all the samples of the training set.
With the loss function, finding optimized $\hat{\theta}$ can be mathematically interpreted as an optimization problem, 
\begin{equation}
	\hat{\theta} = \mathop{\arg\min}_{\theta} \mathcal{L}(\bm{d},\hat{\bm{d}};\theta).
\end{equation} 
Gradient-based optimization methods are usually used to find a local minima of $\mathcal{L}(\bm{d},\hat{\bm{d}};\theta)$. 
Variations of gradient-based methods include the stochastic gradient descent (SGD) 
\cite{robbins1951stochastic} optimizer, Adam optimizer \cite{kingma2015adam}, \emph{etc.}.

\subsection{Embedding Space and Chi-Squared Regression}\label{subsec:embeddingspace}
As mentioned above, the Levenshtein distance has been approximated by 
the $\ell_1$ distance, the Euclidean distance ($\ell_2$ distance), the Hamming distance, \emph{etc.} 
\cite{charikar2006embedding,ostrovsky2007low,andoni2009overcoming,chakraborty2016streaming,zhang2017embedjoin}. 
In this paper, to ensure that gradient-based optimization is applicable, 
the differentiable distances, \emph{i.e.} the $\ell_1$ distance and $\ell_2$ distance, 
are considered as alternative approximations of the Levenshtein distance. 
While in the proposed DSEE, 
the squared Euclidean distance is applied as the approximation of Levenshtein distance. 
Although the squared Euclidean distance does not form a metric space, 
we still find it experimentally and theoretically superior to 
the $\ell_1$ and $\ell_2$ distances 
(see \cref{sec:experiments}, \cref{app:SEbetter} and \cref{app:assumption}). 

Let's make some assumptions or restrictions before further analysis. 
First, given a sequence $\bm{s}$, we assume that each element $u_i$ of the embedding vector $\bm{u}=f(\bm{s})$ 
follows a standard normal distribution $N(0,1)$. 
If the embedding network $f(\cdot)$ uses a batch normalization layer before its output, 
the mean value and standard deviation of $u_i$ are approximately equal to $0$ and $1$ respectively. 
As the normal distribution is the most common type, we expect the distribution of $u_i$ to be close to normal in practice.
Second, if we have trained a good embedding network $f(\cdot)$, 
the elements $u_i$ and $u_j$ are expected to be independent of each other if $i\neq j$. 
This is partly because that the independent embedding elements can hold more information 
about the original sequence. 
Third, given another sequence $\bm{t}$ that is non-homologous to the sequence $\bm{s}$, 
the embedding vector $\bm{v}=f(\bm{t})$ is independent to the embedding vector $\bm{u}$ of sequence $\bm{s}$. 
In addition, the expected distance between the independent embedding vectors $\bm{u}$ and $\bm{v}$ 
should meet the mean value of the Levenshtein distance between two non-homologous sequences. 
To verify whether these assumptions hold in practice, 
we analyze the distributions and independence of the embedding elements in \cref{app:assumption}. 
The results show that the normality assumption is maintained in practice, 
and that the proposed method helps to improve the independence between the embedding elements. 

The average Levenshtein distance between two non-homologous sequences on the DNA-Fountain data is about $80$, 
and we use this number $n=80$ as the dimension of the embedding vectors. 
This setting implies that each position of the two independent embedding vectors 
contributes $1$ to the approximate distance. 
In addition, we also believe that $n=80$ is not too large to cost too much 
computational complexity, nor too small to degrade the performance of the method. 
In order to satisfy the third restriction mentioned above, 
we need to rescale the embedding vector before computing the approximate distance.
The rescaling factor for squared Euclidean distance is 
\begin{equation}
	r_{\ell_2^2} = \frac{\sqrt{2}}{2}.
\end{equation}
For convenience, from now on, the symbols $\bm{u},\bm{v}$ are used to denote the scaled embedding vectors, 
for example, 
\begin{equation}\label{eq:rescaling}
	\bm{u} = r_{\ell_2^2}f(\bm{s})=\frac{\sqrt{2}}{2}f(\bm{s}).
\end{equation}
By rescaling, the vector elements $u_i$,$v_i$ follow the distribution $N(0,1/2)$, 
hence the $u_i-v_i$ follows the standard normal distribution $N(0,1)$ 
when $\bm{u}$ and $\bm{v}$ are independent. 
Recall that the squared Euclidean distance is defined as
\begin{equation}\label{eq:squaredEuclideanDistance}
	d_{\ell_2^2}(\bm{u},\bm{v})= \sum_{i=1}^{n}(u_i-v_i)^2. 
\end{equation}
The squared Euclidean distance between two independent embedding vectors follows 
the chi-squared distribution $\chi^2(n)$ with the embedding dimension $n$ as the degree of freedom. 
As mentioned above, the embedding dimension $n$ is set to $n=80$ to equal 
the average Levenshtein distance of two non-homologous sequences. 
The expected distance between two independent embedding vectors is 
\begin{equation}
	E(d_{\ell_2^2}(\bm{u},\bm{v})) = 80, \quad d_{\ell_2^2}(\bm{u},\bm{v})\sim \chi^2(80), 
\end{equation}
which coincides with the average Levenshtein distance between two non-homologous sequences. 

To train deep models for approximation, 
existing works used the mean squared error, the mean absolute error, 
or their variations as the loss functions on 
the approximate distance $\hat{d}$ and the Levenshtein distance $d$. 
In the gradient descent algorithms, these loss functions and their gradients 
with respect to the predicted distance $\hat{d}$ 
are symmetric about the ground truth distance $d$. 
For example, the mean squared error on $\hat{d}$ and $d$ is 
\begin{equation}
	\mathrm{MSE}(\hat{d},d) = (\hat{d}-d)^2,
\end{equation}
and the partial derivative with respect to $\hat{d}$ 
is $2(\hat{d}-d)$, which are symmetric about $d$. 
However, the distributions of predicted distance $\hat{d}$, the ground truth distance $d$, and 
especially the difference between these two distances $\hat{d}-d$ are skewed rather than symmetric. 
The skewed distributions are more pronounced when the ground truth distance is small. 
For example, a predicted distance $\hat{d} = 2.5$ with positive deviation of $1.5$ 
on ground truth distance $d=1$ is reasonable, 
but in no case should the method predict a distance as $\hat{d} = -0.5$ with negative deviation $-1.5$ on the same ground truth distance. 

To tackle this issue, we introduce the chi-squared regression for training the embedding network. 
Instead of optimizing an approximate error between predicted distance $\hat{d}$ and ground truth distance $d$, 
the chi-squared regression interprets the Levensthein distance $d$ as 
the degree of freedom of $\bm{u}-\bm{v}$ 
and uses a loss function that simulates the relative entropy (also called Kullback–Leibler divergence \cite{kullback1951information}) 
to chi-squared distribution $\chi^2(d)$. 

Recall that the squared Euclidean distance between two independent embedding vectors 
follows the $\chi^2(n)$ distribution with the degree of freedom equal to the embedding dimension. 
For two dependent embedding vectors, the chi-squared distribution is also 
applicable for the distribution of the squared Euclidean distance by the following steps. 
In this paper, 
we call a multivariable $\bm{x} = (x_1,\ldots,x_n)$ to have a degree of freedom $d$, 
iff there exists an orthogonal matrix $\bm{P}$ and a multivariable $\bm{y}$ such that 
\begin{equation}
	\bm{x} = \bm{y}\bm{P} = (y_1,\ldots,y_d,0,\ldots,0)\bm{P}, 
\end{equation}
where the $y_i$s are i.i.d. and follow $N(0,1)$. 
If the difference between two embedding vectors $\bm{u},\bm{v}$
has a degree of freedom $d$, which is
\begin{equation}
	\bm{u}-\bm{v} = \bm{y}\bm{P} = (y_1,\ldots,y_d,0,\ldots,0)\bm{P}, 
\end{equation}
the squared Euclidean distance between them is calculated as 
\begin{align}
	d_{\ell_2^2}(\bm{u},\bm{v}) &= \sum_{i=1}^{n}(u_i-v_i)^2 = (\bm{u}-\bm{v}) (\bm{u}-\bm{v})^T \notag\\
	&= \bm{y}\bm{P}\bm{P}^T\bm{y}^T = \bm{y}\bm{y}^T \notag\\
	&=  \sum_{i=1}^{d}y_i^2,
\end{align}
and follows the chi-squared distribution $\chi^2(d)$ with degree of freedom $d$. 
A step further, the expected value of the squared Eulidean distance between 
$\bm{u}$ and $\bm{v}$ is the degree of freedom $d$ of their difference $\bm{u} - \bm{v}$. 
In view of this, it is a reasonable idea to make connections from 
the Levenshtein distance between the sequences $\bm{s}, \bm{t}$ 
to the degree of freedom of the difference $\bm{u}-\bm{v}$ between their embedding vectors. 
The smaller the Levenshtein distance between the sequences $\bm{s}$ and $\bm{t}$ is,
the more related their embedding vectors $\bm{u}$ and $\bm{v}$ are, 
and the less free variables $y_i$s are needed to support the difference $\bm{u}-\bm{v}$. 
To be precise, if the Levenshtein distance between 
two sequences $\bm{s}$ and $\bm{t}$ is $d$, 
we expect that the difference $\bm{u}-\bm{v}$ between their embedding vectors has 
a degree of freedom $d$. 
By this, the squared Euclidean distance $d_{\ell_2^2}(\bm{u},\bm{v})$ follows 
the $\chi^2(d)$ and its expected value equals the degree of freedom and equals 
the Levenshtein distance between $\bm{s}$ and $\bm{t}$. 

In order to define the loss function between predicted distance $\hat{d}$ to the distribution $\chi^2(d)$, 
the relative entropy is used.
Mathematically, the relative entropy of distrubtion $P$ from $Q$ is defined as 
\begin{equation}
	\mathrm{KLD}(P||Q)=\int_{-\infty}^\infty p(x)\log\frac{p(x)}{q(x)} \mathrm{d} x.
\end{equation}
Let $P$ be the distribution of the predicted distance $\hat{d}$ and $Q_d=\chi^2(d)$ be 
the chi squared distribution with degree of freedom $d$, 
\begin{equation}\label{eq:KLD1}
	\mathrm{KLD}(P||Q_d) = \mathbb{E}_{\hat{d}\sim P}[\log p(\hat{d})] - \mathbb{E}_{\hat{d}\sim P}[\log q_d(\hat{d})].
\end{equation}
For the first term of \cref{eq:KLD1} is not accessable, we use the second term as the optimization target, 
which is 
\begin{align}\label{eq:KLD2}
	\mathrm{RE}\chi^2(\hat{d},d) &=-\log{q_d(\hat{d})}\notag\\
	&=\frac{d}{2} + \log\Gamma\left(\frac{d}{2}\right) - \left(\frac{d}{2}-1\right)\log\hat{d}\notag\\
    &+\frac{\hat{d}}{2}\log\mathrm{e},
\end{align}
where the $\Gamma$ is the Gamma function,the $\mathrm{e}$ is the Euler's number, 
and the $\log$ uses $2$ as the base. 
It is worthy to note that the \cref{eq:KLD2} can also be interpreted as the 
cross-entropy between the distribution of $\hat{d}$ and the chi squared distribution, or a loss 
of negative log-likelihood. 

In summary, the proposed DSEE mainly consists of the following parts:
a deep model $f(\cdot;\theta)$ for mapping the sequences to 
their embedding vectors; 
a Siamese network for calculating the approximations 
of the Levenshtein distance by the squared Euclidean distance; 
and the RE$\chi^2$ loss function for penalizing the difference between 
the approximate and the ground truth distances. 
In the training phase, the parameters $\theta$ are optimized to $\hat{\theta}$ 
subject to minimizing the relative entropy loss RE$\chi^2$, 
while in the testing phase, the trained deep model $f(\cdot;\hat{\theta})$ is used to 
output the embedding vectors of sequences, 
and the squared Euclidean distance between the embedding vectors are used 
as the approximations of the Levenshtein distance.

\section{Experiments and Analysis}\label{sec:experiments} 
The proposed method includes the following variables: 
the embedding network $f(\cdot;\theta)$; 
the embedding space of the Levenshtein distance; 
and the loss function to be optimized. 
To find suitable network structures for the embedding network $f(\cdot;\theta)$, 
several mainstream structures, 
the CNN-ED-5, the CNN-ED-10, the RNN, and the GRU, 
are applied in the experiments. 
To show the advantages of the proposed squared Euclidean embedding, 
the experiments on the $\ell_1$ embedding and the $\ell_2$ embedding 
of the Levenshtein distance are conducted. 
Finally, to illustrate the advantages of the proposed chi-square regression, 
we conduct comparative experiments using the mean squared error and mean absolute error as 
loss functions. 

To obtain a comprehensive view on how these three variables interact 
with each other and affect the performance of the proposed method, 
all combinations of these experimental setups are explored. 
The results are reported in \cref{tab:experiments}.
In this table, the column headers indicate the engaged 
structures of the embedding network and the loss functions used in the training phase, 
for example, the ``CNN-ED-5: RE$\chi^2$'' means the CNN-ED-5 structure and the RE$\chi^2$ loss.
The row headers indicate the metrics used to evaluate the testing performance of the methods
and the embedding spaces of the features, 
for example, the ``$\mathrm{AE}_h$: $\ell_2$'' means the $\mathrm{AE}_h$ metric predifined in \cref{subsec:metrics} and 
the $\ell_2$ embedding of the sequences. 
The results in \cref{tab:experiments} are reported 
in the format ``$\mathrm{mean}\pm\mathrm{std}$'' of the mean value 
and the standard deviation over $5$ runs of the experiments. 
It is worth noting that the RE$\chi^2$ loss is incompatible with the $\ell_1$ and $\ell_2$ 
embeddings of the sequences, and we make the relevant numbers italicized in \cref{tab:experiments}. 
We have also marked the ``good'' results in boldface, 
which are the $\mathrm{AE}_h$ less than $1.00$ 
and the $\mathrm{OA}$ greater than $99.90\%$.

\begin{table*}[htb]
	\caption{Results of the experiments.
	The column headers indicate the engaged 
	structures of the embedding network and the loss functions used in the training phase,
	while the row headers indicate the metrics used to evaluate the testing performance 
	and the embedding spaces of the features. 
	The results are reported in the format ``$\mathrm{mean}\pm\mathrm{std}$'' of the mean value 
	and the standard deviation over $5$ runs of the experiments. 
	The italicized numbers are the results obtained by applying RE$\chi^2$ loss on the $\ell_1$ and $\ell_2$ embeddings, 
	where the RE$\chi^2$ loss is incompatible with this embedding spaces. 
	The boldface numbers are the ``good'' results of $\mathrm{AE}_h$ less than $1.00$ 
	and the $\mathrm{OA}$ greater than $99.90\%$.}\label{tab:experiments}
	\vskip 0.15in
    \begin{minipage}{\linewidth}
		\resizebox{\linewidth}{!}{
		\begin{tabular}{x{0.06\linewidth}x{0.06\linewidth}x{0.127\linewidth}x{0.127\linewidth}x{0.127\linewidth}x{0.127\linewidth}x{0.127\linewidth}x{0.127\linewidth}}
			\toprule
            \multirow{2}{*}{Metric}  & \multirow{2}{*}{Embed} & \multicolumn{3}{c}{CNN-ED-5}                                                           & \multicolumn{3}{c}{CNN-ED-10}                                                          \\ \cmidrule(lr){3-5}\cmidrule(lr){6-8} 
										&                         & \multicolumn{1}{c}{MSE} & \multicolumn{1}{c}{MAE} & \multicolumn{1}{c}{RE$\chi^2$} & \multicolumn{1}{c}{MSE} & \multicolumn{1}{c}{MAE} & \multicolumn{1}{c}{RE$\chi^2$}   \\ \midrule
										& $\ell_1$           & 4.74 $\pm$ 0.03         & 3.57 $\pm$ 0.18         & \emph{5.66 $\pm$ 0.15}         & 4.60 $\pm$ 0.13         & 3.70 $\pm$ 0.16         & \emph{5.26 $\pm$ 0.04}         \\
										& $\ell_2$           & 6.23 $\pm$ 0.01         & 3.73 $\pm$ 0.11         & \emph{6.02 $\pm$ 0.42}         & 5.93 $\pm$ 0.07         & 3.56 $\pm$ 0.06         & \emph{5.00 $\pm$ 0.06}         \\
		\multirow{-3}{*}{$\mathrm{AE}$} & $\ell_2^2$           & 4.20 $\pm$ 0.06         & 4.12 $\pm$ 0.02         & 4.67 $\pm$ 0.09                  & 4.11 $\pm$ 0.01         & 4.12 $\pm$ 0.08         & 4.53 $\pm$ 0.03                  \\\midrule
										& $\ell_1$         & 3.50 $\pm$ 0.02         & 2.50 $\pm$ 0.15         & \emph{1.89 $\pm$ 0.05}         & 3.44 $\pm$ 0.04         & 2.71 $\pm$ 0.19         & \emph{1.96 $\pm$ 0.01}         \\
										& $\ell_2$         & 5.99 $\pm$ 0.01         & 2.69 $\pm$ 0.08         & \emph{2.59 $\pm$ 0.03}         & 5.88 $\pm$ 0.02         & 2.69 $\pm$ 0.14         & \emph{2.77 $\pm$ 0.05}         \\
		\multirow{-3}{*}{$\mathrm{AE}_h$}& $\ell_2^2$         & \textbf{0.90 $\pm$ 0.05}& \textbf{0.96 $\pm$ 0.09}&\textbf{0.90 $\pm$ 0.00}          & 1.11 $\pm$ 0.02         & 1.56 $\pm$ 0.15         & \textbf{0.91 $\pm$ 0.01}         \\\midrule
										& $\ell_1$           &\textbf{99.98 $\pm$ 0.00}& 96.57 $\pm$ 0.30        & \emph{99.42 $\pm$ 0.11}        &\textbf{99.98 $\pm$ 0.01}& 96.59 $\pm$ 0.27        & \emph{99.27 $\pm$ 0.04}        \\
										& $\ell_2$           & 99.85 $\pm$ 0.01        & 96.40 $\pm$ 0.26        & \emph{98.34 $\pm$ 0.09}        & 99.66 $\pm$ 0.06        & 96.81 $\pm$ 0.09        & \emph{98.14 $\pm$ 0.02}        \\
		\multirow{-3}{*}{$\mathrm{OA}$} & $\ell_2^2$           &\textbf{99.98 $\pm$ 0.01}&\textbf{99.85 $\pm$ 0.08}&\textbf{99.98 $\pm$ 0.00}         &\textbf{99.91 $\pm$ 0.00}&99.06 $\pm$ 0.16         & \textbf{99.98 $\pm$ 0.01}        \\
			\bottomrule
        \end{tabular}
		}
	\end{minipage}\\[12pt]
	\begin{minipage}{\linewidth}
		\resizebox{\linewidth}{!}{
		\begin{tabular}{x{0.06\linewidth}x{0.06\linewidth}x{0.127\linewidth}x{0.127\linewidth}x{0.127\linewidth}x{0.127\linewidth}x{0.127\linewidth}x{0.127\linewidth}}
            \toprule
            \multirow{2}{*}{Metric}  & \multirow{2}{*}{Embed}     & \multicolumn{3}{c}{RNN}                                                                & \multicolumn{3}{c}{GRU}                                                    \\ \cmidrule(lr){3-5}\cmidrule(lr){6-8} 
										&                         & \multicolumn{1}{c}{MSE} & \multicolumn{1}{c}{MAE} & \multicolumn{1}{c}{RE$\chi^2$} & \multicolumn{1}{c}{MSE} & \multicolumn{1}{c}{MAE} & \multicolumn{1}{c}{RE$\chi^2$} \\ \midrule
										& $\ell_1$           & 5.25 $\pm$ 0.05         & 4.32 $\pm$ 0.43         & \emph{5.89 $\pm$ 0.18}         & 4.61 $\pm$ 0.14         & 3.45 $\pm$ 0.26         & \emph{5.36 $\pm$ 0.06}  \\
										& $\ell_2$           & 7.15 $\pm$ 0.08         & 5.11 $\pm$ 0.44         & \emph{6.71 $\pm$ 0.33}         & 7.52 $\pm$ 0.15         & 3.89 $\pm$ 0.15         & \emph{5.32 $\pm$ 0.05}  \\
		\multirow{-3}{*}{$\mathrm{AE}$}	& $\ell_2^2$           & 4.31 $\pm$ 0.01         & 4.36 $\pm$ 0.06         & 5.41 $\pm$ 0.02                  & 3.98 $\pm$ 0.02         & 4.05 $\pm$ 0.02         & 5.51 $\pm$ 0.05  \\\midrule
										& $\ell_1$         & 4.06 $\pm$ 0.05         & 3.25 $\pm$ 0.28         & \emph{2.25 $\pm$ 0.03}         & 3.55 $\pm$ 0.09         & 2.48 $\pm$ 0.27         & \emph{2.09 $\pm$ 0.05}  \\
										& $\ell_2$         & 6.49 $\pm$ 0.15         & 3.56 $\pm$ 0.26         & \emph{3.15 $\pm$ 0.14}         & 6.40 $\pm$ 0.04         & 2.75 $\pm$ 0.09         & \emph{2.73 $\pm$ 0.02}  \\
		\multirow{-3}{*}{$\mathrm{AE}_h$}& $\ell_2^2$         & 1.03 $\pm$ 0.02         & 1.14 $\pm$ 0.24         & \textbf{0.91 $\pm$ 0.01}         & \textbf{0.73 $\pm$ 0.03}& \textbf{0.67 $\pm$ 0.02}& \textbf{0.88 $\pm$ 0.00}  \\\midrule
										& $\ell_1$           &\textbf{99.96 $\pm$ 0.01}& 96.51 $\pm$ 0.42        & \emph{99.15 $\pm$ 0.13}        &\textbf{99.98 $\pm$ 0.00}& 96.72 $\pm$ 0.25        & \emph{99.40 $\pm$ 0.14} \\
										& $\ell_2$           & 99.77 $\pm$ 0.02        & 96.10 $\pm$ 0.87        & \emph{98.23 $\pm$ 0.13}        & 99.88 $\pm$ 0.01        & 96.25 $\pm$ 0.26        & \emph{98.21 $\pm$ 0.02} \\
		\multirow{-3}{*}{$\mathrm{OA}$}	& $\ell_2^2$           &\textbf{99.96 $\pm$ 0.00}& 99.77 $\pm$ 0.18        & \textbf{99.94 $\pm$ 0.00}        &\textbf{100.00$\pm$ 0.00}&\textbf{99.99 $\pm$ 0.00}& \textbf{99.97 $\pm$ 0.00} \\
            \bottomrule
        \end{tabular}
		}
	\end{minipage}
\vskip -0.1in
\end{table*}

As stated in \cref{subsec:metrics}, 
the metric $\mathrm{AE}$ presents the global approximation error for all the homologous and 
non-homologous sequence pairs. 
A small $\mathrm{AE}$ indicates that the model is well trained and performs well. 
However, a precise approximation of the Levenshtein distance for non-homologous pair is 
less of our interest, hence under the premise that $\mathrm{AE}$ is small, the 
$\mathrm{AE}_h$ is a more important metric to evaluate the testing performance of the models. 
As illustrated in \cref{tab:experiments}, all the experiments show small 
$\mathrm{AE}$, in view of this, 
the $\mathrm{AE}_h$ and the $\mathrm{OA}$ will be discussed mainly in further analysis. 

The squared Euclidean embedding takes the lead by a large margin. 
Let us fix the structure of embedding network and loss function, and 
compare the embedding spaces. 
The squared Euclidean embedding is several times superior to the $\ell_1$ embedding 
in metric $\mathrm{AE}_h$ over all the combinations of embedding networks and loss functions. 
$\mathrm{OA}$ of squared Euclidean embedding is also higher than $\mathrm{OA}$ of $\ell_1$ embedding 
in most cases. 
Meanwhile, the $\ell_2$ embedding performs the worst among three embedding spaces. 
Briefly, the three embeddings are ranked as $\ell_2^2>\ell_1>\ell_2$. 
We provide a preliminary analysis of 
these results in \cref{app:SEbetter}. 

The proposed chi-squared regression also shows good performance. 
When the squared Euclidean embedding is used, applying the RE$\chi^2$ loss
improves the performance of the networks CNN-ED-10 and RNN in 
metrics $\mathrm{AE}_h$ and $\mathrm{OA}$ by a large margin, 
while on the networks CNN-ED-5 and GRU, the RE$\chi^2$ loss also has comparable 
performance to MAE and MSE losses. 
As mentioned above, 
the RE$\chi^2$ loss is incompatible with the $\ell_1$ and $\ell_2$ embeddings 
under the assumptions in \cref{subsec:embeddingspace}. 
However, brutely applying the RE$\chi^2$ loss on the $\ell_1$ and $\ell_2$ embeddings 
still shows performance better than or equal to the MSE and MAE losses.
We conjecture that the powerful fitting ability of neural networks overcomes the 
inability of the $\ell_1$ and $\ell_2$ embeddings to conduct the 
chi-squared distribution $\chi^2(d)$. 

As illustrated in \cref{tab:experiments}, 
different choices of the embedding network $f(\cdot;\theta)$ have similar performance for most 
combinations of embedding spaces and loss functions. 
The only exception occurs on squared Euclidean embedding, to be precise, 
GRU outperforms the other three embedding networks in metric $\mathrm{AE}_h$, 
no matter which loss function is engaged. 
Because GRU is the most modern and complex one among four models, 
it was believed to have better performance globally. 
However, experiments have shown that only the squared Euclidean embedding enables the GRU. 
We may infer that the application of the squared Euclidean embedding not only helps to improve the 
performance, but also helps to discover the potential of complex embedding networks. 
It is also worth noting that, no matter which model is used, the combination 
of squared Euclidean embedding and the RE$\chi^2$ loss always 
has a stable and excellent performance. 

We leave the details of the experiments to the Appendices, 
including the rescaling factors for the $\ell_1$ and $\ell_2$ embeddings \cref{app:resalingfactor}
and the setups for applied embedding networks \cref{app:networks}. 

\section{Conclusion}\label{sec:conclusion}
Fast approximation of the Levenshtein distance is demanded by a lot applications. 
In this paper, raised from DNA storage researches, 
the deep squared Euclidean embedding (DSEE) of DNA sequences was proposed 
to approximate the Levenshtein distance. 
The proposed method consists of three main components, 
namely, the Siamese neural network, the squared Euclidean embedding, 
and the chi-squared regression. 
It is innovative in applying the squared Euclidean embedding 
to approximate the Levenshtein distance, 
interpreting the Levenshtein distance between the sequences 
as the degree of freedom of their embeddings,
and introducing the relative entropy to $\chi^2$ distribution as the loss function. 
The advantages of the proposed DSEE were theoretically analyzed under 
concise and reasonable assumptions. 
We also conducted comprehensive experiments and ablation studies. 
The experimental results echoed the theoretical analysis and showed that 
the proposed DSEE is powerful in approximating the Levenshtein distance.

\section*{Acknowledgements}
The authors are grateful to the reviewers for their valuable comments. 
This work was supported by the National Key Research and Development Program of China under Grant 2020YFA0712100 
and the National Natural Science Foundation of China.

\bibliography{AlanGuo-bio}
\bibliographystyle{icml2022}

\newpage
\appendix
\section{Rescaling factors for $\ell_1$ and $\ell_2$ embeddings}\label{app:resalingfactor}
The $\ell_1$ and $\ell_2$ distances between the embedding vectors 
are considered as alternative approximations of the Levenshtein distance 
between the original sequences. 
To satisfy the restriction that the approximate distance between two independent 
embedding vectors should equal to the average Levenshtein distance between two non-homologous sequences, 
the $\ell_1$ and $\ell_2$ embeddings also need to be rescaled. 

The rescaling factor for $\ell_1$ embedding vectors is 
\begin{equation}
	r_{\ell_1}=\frac{\sqrt{\pi}}{2}.
\end{equation} 
Given two independent embedding vectors $\bm{u}$ and $\bm{v}$, the $\ell_1$ distance between them is
\begin{equation}\label{eqn:ell1}
	d_{\ell_1}(\bm{u},\bm{v})= \sum_{i=1}^{n}|u_i-v_i|. 
\end{equation} 
By rescaling, each element of the vectors $\bm{u}$ and $\bm{v}$ follows the normal distribution 
$N(0,\pi/4)$ independently. 
Therefore, the $|u_i-v_i|$ in \cref{eqn:ell1} follows the half-normal distribution $HN(\sqrt{\pi/2})$, 
and the expected value of $|u_i-v_i|$ is $E(|u_i-v_i|) = 1$\footnote{The expected value of 
$x,\, x\sim HN(\sigma)$ is $E(x)=\frac{\sigma\sqrt{2}}{\sqrt{\pi}}$.}.
The expected value of $d_{\ell_1}(\bm{u},\bm{v})$ in \cref{eqn:ell1} is easily computed 
as the dimension of the embedding vector
\begin{equation}
	E(d_{\ell_1}(\bm{u},\bm{v}))= n = 80,
\end{equation} 
which is also the average Levenshtein distance between two non-homologous sequences. 
It is worth noting that the distribution of $d_{\ell_1}(\bm{u},\bm{v})$ is not easy to depict, 
and the relative entropy to these family of distributions is difficult to calculate. 

The rescaling factor for $\ell_2$ embedding vectors is 
\begin{equation}
	r_{\ell_2} = \frac{n \Gamma(n/2)}{2\Gamma((n+1)/2)}.
\end{equation}
The $\ell_2$ distance between two independent embedding vectors $\bm{u}$ and $\bm{v}$ is calculated by
\begin{equation}\label{eqn:ell2}
	d_{\ell_2}(\bm{u},\bm{v})= \left(\sum_{i=1}^{n}(u_i-v_i)^2\right)^{\frac{1}{2}}. 
\end{equation}
In view of the independence of the vector elements of $\bm{u}$ and $\bm{v}$, 
the $u_i-v_i$ follows the normal distribution, which is 
\begin{equation}
	\frac{u_i-v_i}{\sqrt{2}r_{\ell_2}} \sim N(0,1).
\end{equation}
Hence, the rescaled $\ell_2$ distance between $\bm{u}$ and $\bm{v}$
\begin{equation}
	\frac{d_{\ell_2}(\bm{u},\bm{v})}{\sqrt{2}r_{\ell_2}} = \left(\sum_{i=1}^{n}\left(\frac{u_i-v_i}{\sqrt{2}r_{\ell_2}}\right)^2\right)^{\frac{1}{2}}, 
\end{equation}
follows the chi distribution $\chi(n)$ with degree $n$ of freedom. 
By the formula of expected value of chi distribution\footnote{
The expected value of 
$x,\, x\sim \chi(n)$ is $E(x)=\frac{\sqrt{2}\Gamma((n+1)/2)}{\Gamma(n/2)}$.
}, the expected value of the $\ell_2$ distance is 
\begin{align}
	E(d_{\ell_2}(\bm{u},\bm{v})) &= \sqrt{2}r_{\ell_2} E\left(\frac{d_{\ell_2}(\bm{u},\bm{v})}{\sqrt{2}r_{\ell_2}}\right)\notag\\
	&= \sqrt{2}\frac{n \Gamma(n/2)}{2\Gamma((n+1/2))}\cdot\sqrt{2} \frac{\Gamma((n+1)/2)}{\Gamma(n/2)}\notag\\
	&= n,
\end{align}
where the $n=80$ is the preset dimension of the embedding vector and the average Levenshtein distance between two non-homologous sequences. 
Similar to the $\ell_1$ embedding, the $\ell_2$ embedding is also incompatible with the relative entropy. 
Given a chi distribution $\chi(k)$, the expected value is 
\begin{equation}
	E(x) = \sqrt{2}\frac{\Gamma((k+1)/2)}{\Gamma(k/2)},\quad x\sim\chi(k),
\end{equation}
which does not equal the parameter $k$ of the chi distribution. 
If we apply the ground truth Levenshtein distance $d$ 
as the degree of freedom of the chi distribution 
and force the predicted $\ell_2$ distance $\hat{d}$ to obey $\chi(d)$, 
the expected value of $\hat{d}$ does not equal to the ground truth $d$. 
In view of this, the relative entropy to chi distribution is defined meaningless. 

\section{Details on the embedding networks}\label{app:networks}
In the experiments, four structures of the embedding networks are used, namely the 
CNN-ED-5, the CNN-ED-10, the RNN, and the GRU. 

The CNN-ED-5 and CNN-ED-10 were proposed in \cite{dai2020convolutional}. 
These two CNNs have similar structure by stacking layers of CNNs, activations, and average poolings. 
Taking the CNN-ED-5 as an example, the model stacks the following layers in five times, 
which are the $1$D-CNN (output channels: $64$, 
kernel size: $3$, stride: $1$, and padding size: $1$), 
the average pooling (kernel size: $2$), 
and the activation of ReLU. 
After these CNNs, two cascaded fully connected layers transform the features into a dimension of $n=80$, 
and a batch normalization is engaged to force the output to follow the $N(0,1)$. 
The CNN-ED-10 doubles the $1$D-CNN layers in CNN-ED-5, 
while the other parts of the structure remain the same. 
The model RNN is a stacked RNN of two recurrent units and bidirectional. 
The size of its hidden features is $64$ and the activation is $\mathrm{tanh}$. 
After the recurrent units, the model of RNN uses the same top of the fully connected layers and batch normalization with 
CNN-ED-X. 
The model GRU totally follows the same structure of model RNN, but it uses the 
GRU as the recurrent units. 
The parameters of its recurrent units are also the same with the model RNN. 

\section{Why $\ell_2^2>\ell_1>\ell_2$?}\label{app:SEbetter}
The chi-squared regression is based on the straightforward assumption 
that the degree of freedom of the difference between the embedding vectors 
is consistent with the Levenshtein distance. 
To be precise, let $\bm{s},\bm{t}$ be two random sequences with Levenshtein distance 
$d$, and $\bm{u},\bm{v}$ be their respective embedding vectors, 
there exists an orthogonal matrix $\bm{P}$ such that 
\begin{align}
	\bm{u}-\bm{v} &= \bm{y}\bm{P}\notag\\
	&= (y_1,y_2,\ldots,y_d,0,\ldots,0)\bm{P}\label{app:eqn:degreeoffreedom}, 
\end{align}
where the $y_i$ are random variables that follow the standard normal distribution 
$N(0,1)$ independently, and the $d$ is the Levenshtein distance $d_L(\bm{s},\bm{t})$ 
between $\bm{s}$ and $\bm{t}$. 

Under this assumption, we demonstrate that the expected value of $\ell_2^2$ distance 
between two embedding vectors $\bm{u},\bm{v}$ matches 
the degree of freedom $d$ of $\bm{u}-\bm{v}$, 
which is also the Levenshtein distance $d$ between sequences $\bm{s}$ and $\bm{t}$. 
But the expected values of $\ell_1$ and $\ell_2$ distances are not consistent 
with the degree of freedom of $\bm{u}-\bm{v}$. 
This suggests that the $\ell_2^2$ distance is preferred over the $\ell_1$ and $\ell_2$ distances.
For convenience, we use the notation $\bm{x} = \bm{u}-\bm{v}$ in the following text. 

By \cref{app:eqn:degreeoffreedom}, the squared Euclidean distance is calculated as 
\begin{align}
	d_{\ell_2^2}(\bm{u},\bm{v})&= \bm{x}\bm{x}^T=\bm{y}\bm{P} \bm{P}^T\bm{y}^T = \bm{y}\bm{y}^T \notag \\
	&= \sum_{i=1}^{d} y_i^2, 
\end{align}
and follows the chi-squared distribution $\chi^2(d)$ with degree $d$ of freedom. 
The $\ell_2$ distance is the square root of the squared Euclidean distance, 
so it follows the chi distribution $\chi(d)$ with degree $d$ of freedom. 
As for the $\ell_1$ distance, we know that if the orthogonal matrix $\bm{P}$ is 
a signed permutation matrix, \emph{i.e.} orthogonal matrix in $\{0,1,-1\}$, the 
$\ell_1$ distance
\begin{equation}
	d_{\ell_1}(\bm{u},\bm{v}) = \sum_{i=1}^{n}|x_i| = \sum_{i=1}^{d}|y_i|
\end{equation}
is the summation of independent half-normal distributions $HN(1)$ in $d$ times. 
However, when $\bm{P}$ is an arbitrary orthogonal matrix, 
the distribution followed by the $\ell_1$ distance is difficult to compute. 
We use the notation $Q(d,\bm{P})$ to denote the unknown distribution 
of $\ell_1$ distance in the following text. 

\begin{figure}[htb]
	\vskip 0.2in
	\begin{center}
	\centerline{\includegraphics[width=0.9\linewidth]{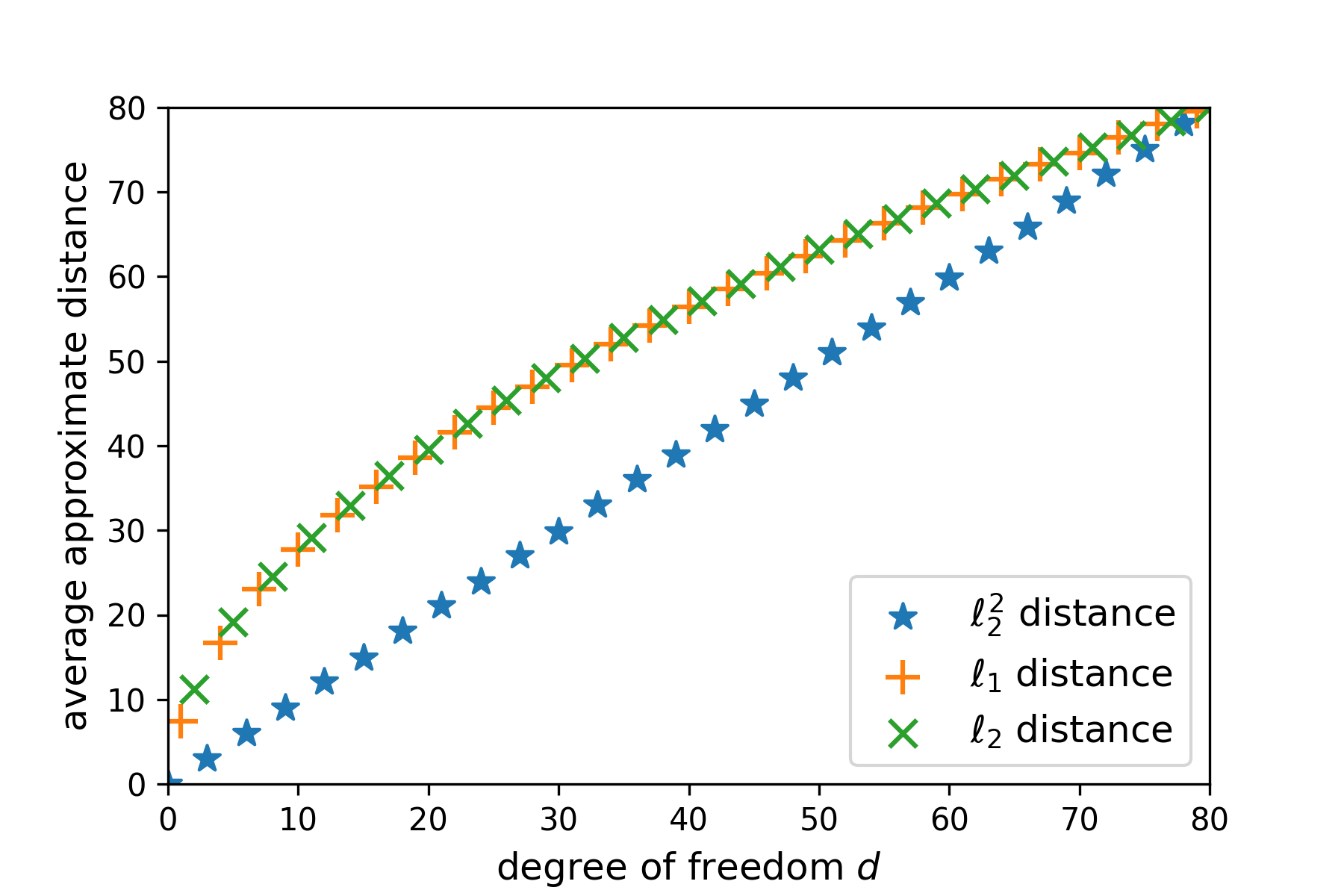}}
	\caption{The average approximate distances with respect to the degree of freedom, from Monte Carlo simulations. 
	The average squared Euclidean distance is linear to the degree of freedom, while the 
	average $\ell_1$ and $\ell_2$ distances are not.}
	\label{fig:sim1}
	\end{center}
	\vskip -0.2in
\end{figure}

The expected value of the squared Euclidean distance is 
\begin{equation}
	E(d_{\ell_2^2}) = d, \, d_{\ell_2^2}\sim\chi^2(d),
\end{equation}
and the expected value of the $\ell_2$ distance is 
\begin{equation}\label{app:eqn:ell2}
	E(d_{\ell_2}) = \frac{\sqrt{2}\Gamma((d+1)/2)}{\Gamma(d/2)}, \, d_{\ell_2}\sim\chi(d),
\end{equation}
while the expected value of the $\ell_1$ distance is complex 
and left to the Monte Carlo simulation. 
It is easy to figure out that the expected value of the $\ell_2$ distance 
in \cref{app:eqn:ell2} is not equal to the degree of freedom $d$. 
In \cref{fig:sim1}, we plot the average approximate distances from Monte Carlo simulations 
respect to different degrees of freedom. 
Although we have rescaled the $\bm{x}$ so that the average approximate distance at $d=80$ 
is equal to $d$, 
the average $\ell_1$ and $\ell_2$ distances are still not equal to $d$ when $d\neq 80$, 
because of the non-linear relationship between their expected values and $d$. 
In this view, the squared Euclidean distance is superior to 
the $\ell_1$ and $\ell_2$ distances under the above assumption. 

\begin{figure}[htb]
	\vskip 0.2in
	\begin{center}
	\centerline{\includegraphics[width=0.9\linewidth]{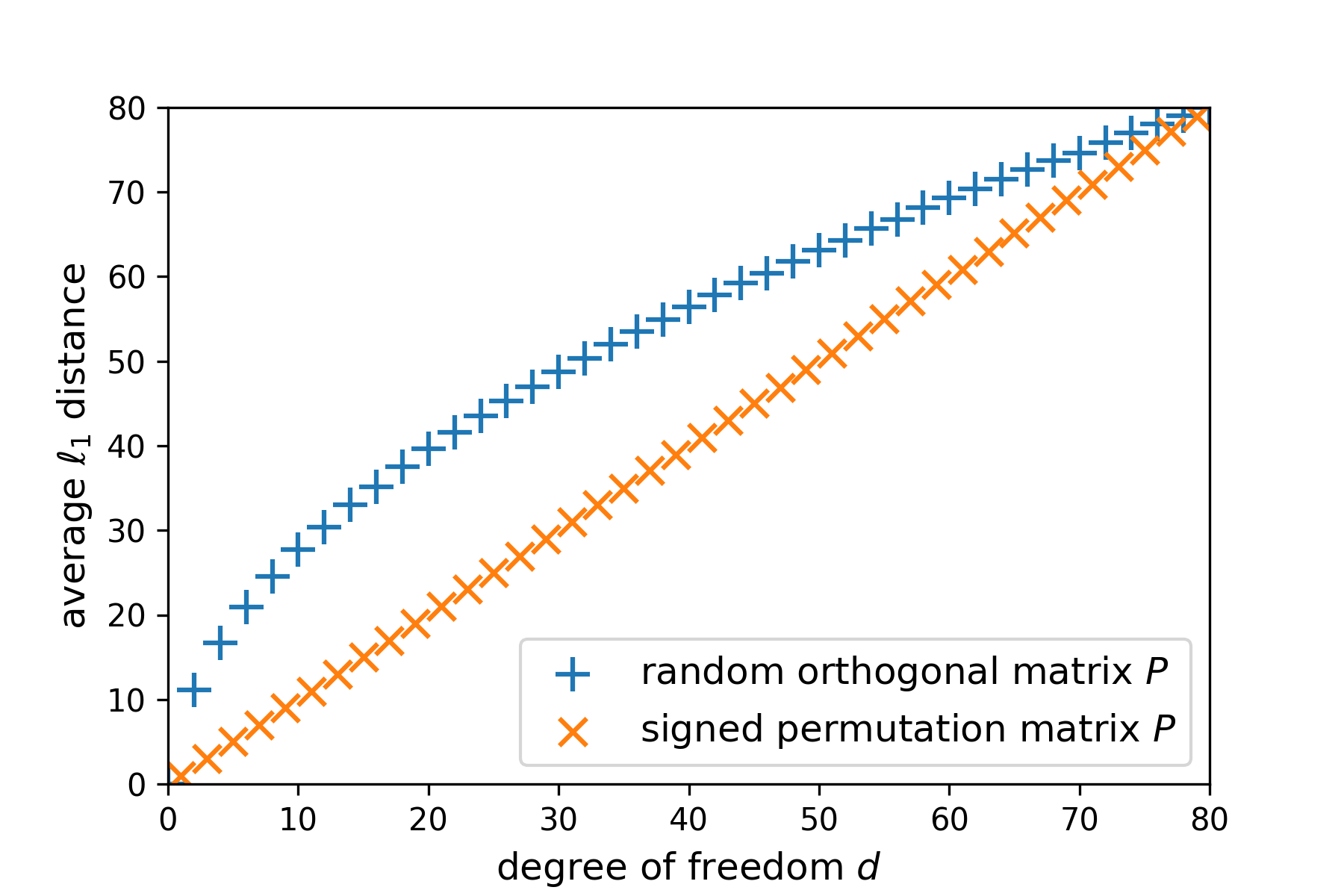}}
	\caption{The average $\ell_1$ distances with respect to the degree of freedom under 
	different choices of orthogonal matrix $\bm{P}$, from Monte Carlo simulations. 
	The average $\ell_1$ distances with a signed permutation matrix $\bm{P}$ is linear to the degree of freedom, 
	while the average $\ell_1$ distance with random orthogonal matrix $\bm{P}$ is not.}
	\label{fig:sim2} 
	\end{center}
	\vskip -0.2in 
\end{figure}

From \cref{fig:sim1}, 
it is straightforward to infer that the $\ell_1$ distance should perform 
as poorly as the $\ell_2$ distance. 
However, the experimental results in \cref{tab:experiments} show not only that the $\ell_2^2$ is the best among three, 
but also that the $\ell_1$ distance performs better than $\ell_2$ distance. 
A restriction on the orthogonal matrix $\bm{P}$ may answer this question. 
Recall that the distribution $Q(d,\bm{P})$ of the $\ell_1$ distance is related 
to both the orthogonal matrix $\bm{P}$ and the degree of freedom $d$. 
When the orthogonal matrix $\bm{P}$ is restricted to a signed permutation matrix, 
the distribution $Q(d,\bm{P})$ is a summation of the half-normal distributions $HN(1)$, 
and the expected value of the $\ell_1$ distance becomes linear with the degree $d$. 
The Monte Carlo simulation results are plotted in \cref{fig:sim2} to illustrate the 
relationship between the average distances and the orthogonal matrix $\bm{P}$. 
From this figure, it can be confirmed that 
the $\ell_1$ distance has the potential to return an expected value equal to the 
degree of freedom.  
We speculate that the neural networks in the experiments of $\ell_1$ embedding 
learned that the latent matrix $\bm{P}$ obeys some restrictions, 
such as $\bm{P}$ as the signed permutation matrix, 
and lead to better performance compare to the $\ell_2$ distance. 

In summary, the theoretical results suggest that applying the squared Euclidean  
distance outperforms both $\ell_1$ and $\ell_2$ distances, 
and the $\ell_1$ distance is better than the $\ell_2$ distance. 
This is consistent with the experimental results in \cref{tab:experiments}. 

\section{Do these assumptions hold?}\label{app:assumption}
In \cref{subsec:embeddingspace}, the following two of the three assumptions about the embedding vectors are mostly concerned 
in practice.
\begin{itemize}
	\item The element $u_i$ of the embedding vector $\bm{u}$ is a random variable 
	with a standard normal distribution $N(0,1)$. 
	\item The elements $u_i,u_j$ of the embedding vector $\bm{u}$ are expected to be independent of each other 
	if $i\neq j$. 
\end{itemize}

\begin{figure*}[htb]
	\vskip 0.2in
	\begin{center}
	\centerline{\includegraphics[width=0.9\linewidth]{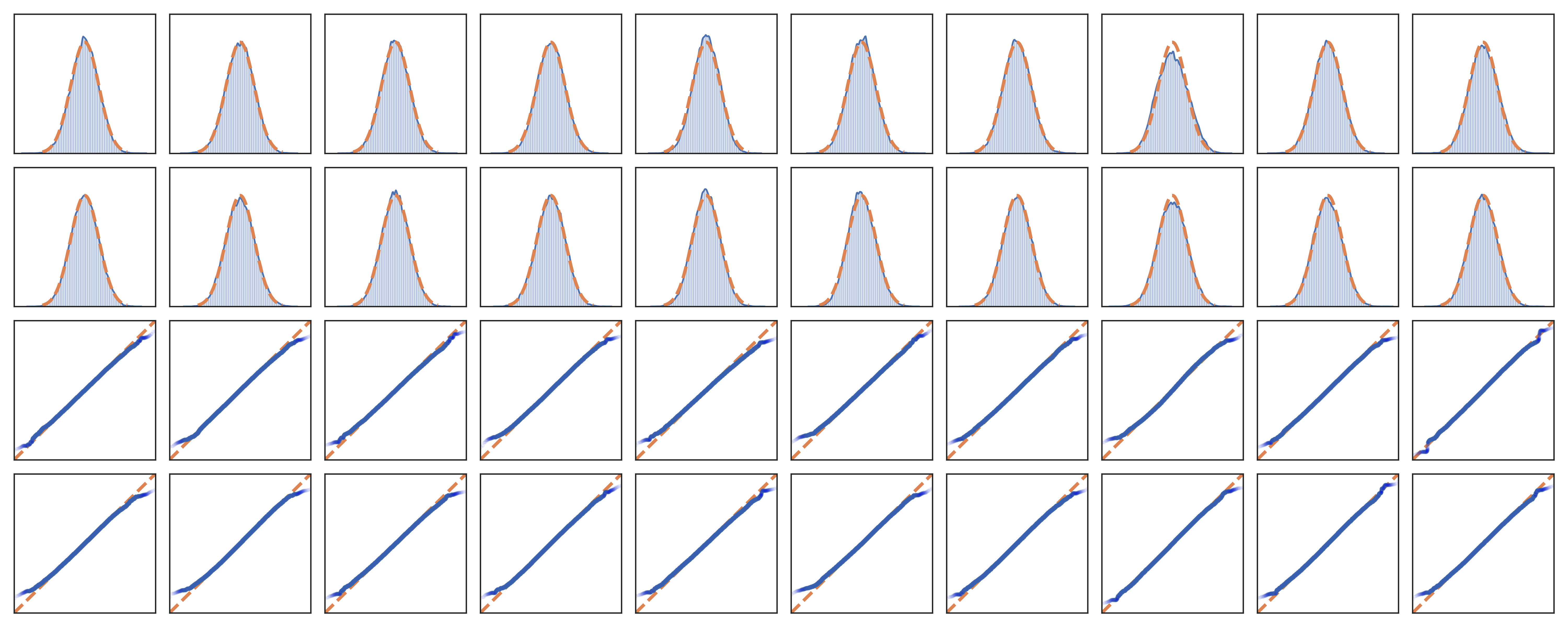}}
	\caption{Distributions of the first $20$ elements of testing embedding vectors 
	compared to $N(0,1)$. The upper subfigures are the estimated PDFs of the distributions of $u_i$s. 
	The lower subfigures are the QQ plots of the $u_i$s. 
	The orange dashed curves and lines are the PDF of $N(0,1)$ and the reference line to $N(0,1)$, 
	respectively. }
	\label{fig:distribution} 
	\end{center}
	\vskip -0.2in  
\end{figure*}

As mentioned in \cref{subsec:embeddingspace}, the batch normalization is applied to the embedding vectors. 
The $\mathrm{BatchNorm}$ layer maintains the embedding element $u_i$ with  
$\mathrm{mean}\approx 0,\mathrm{std}\approx1$ by updating the mean and variance per training batch 
with a momentum term. 
If the testing sample uses the same distribution as the training sample, the testing 
distribution for $u_i$ will also have $\mathrm{mean}\approx 0,\mathrm{std}\approx1$. 
Since the normal distribution is the most common type, we speculate that the distribution 
of $u_i$ is close to the standard normal distribution $N(0,1)$ and no additional regularization is required.
To verify our conjecture, we plot the distributions of the first $20$ elements of 
the embedding vector $\bm{u}$ from the testing phase in \cref{fig:distribution}. 
One could see that the distribution of each $u_i$ is close to $N(0,1)$, 
which suggests that the first assumption is reasonable and fits well with the real-world data. 

\begin{figure*}[htb]
		\vskip 0.2in
		\begin{center}
			\includegraphics[height=0.23\linewidth]{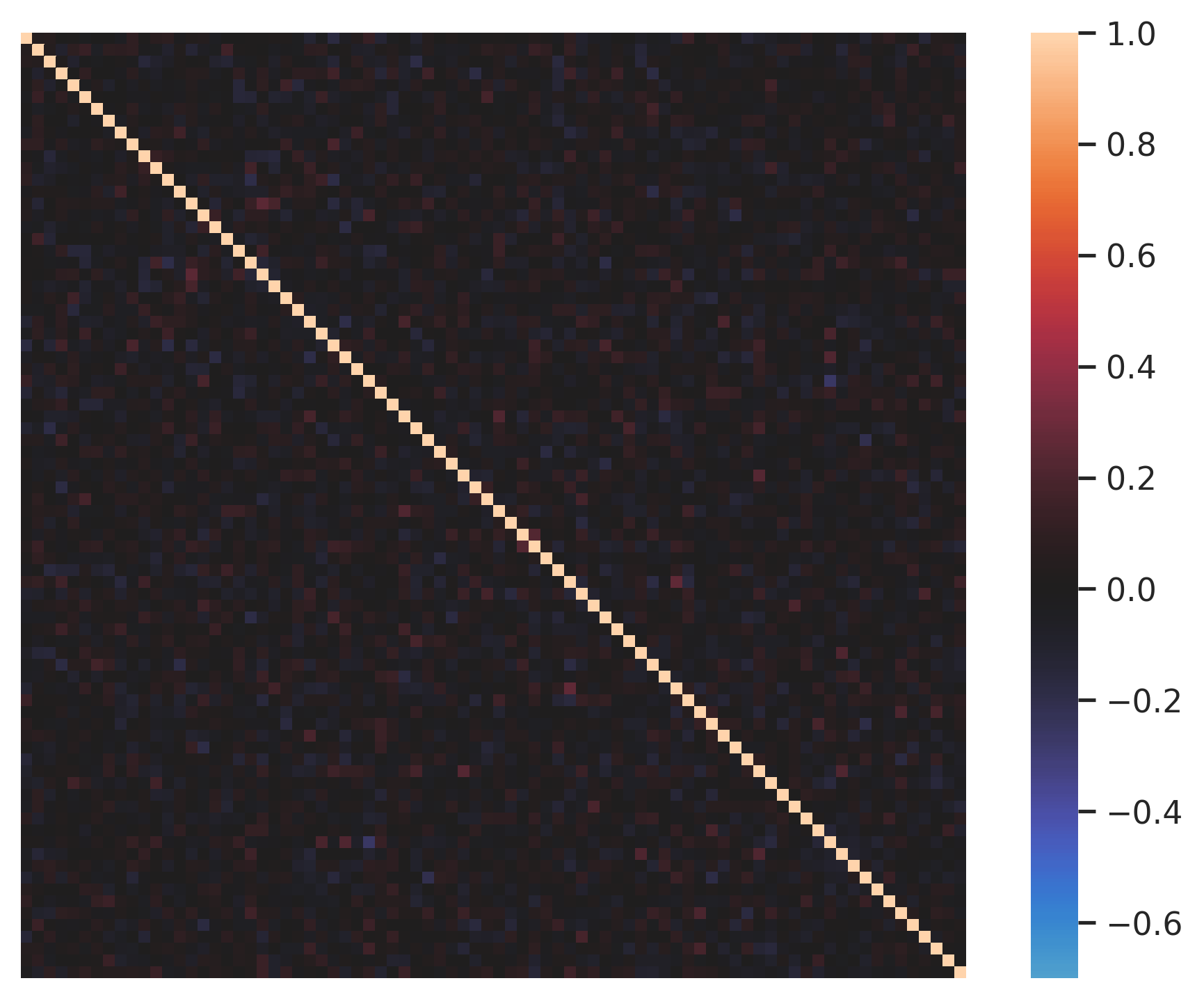}
			\includegraphics[height=0.23\linewidth]{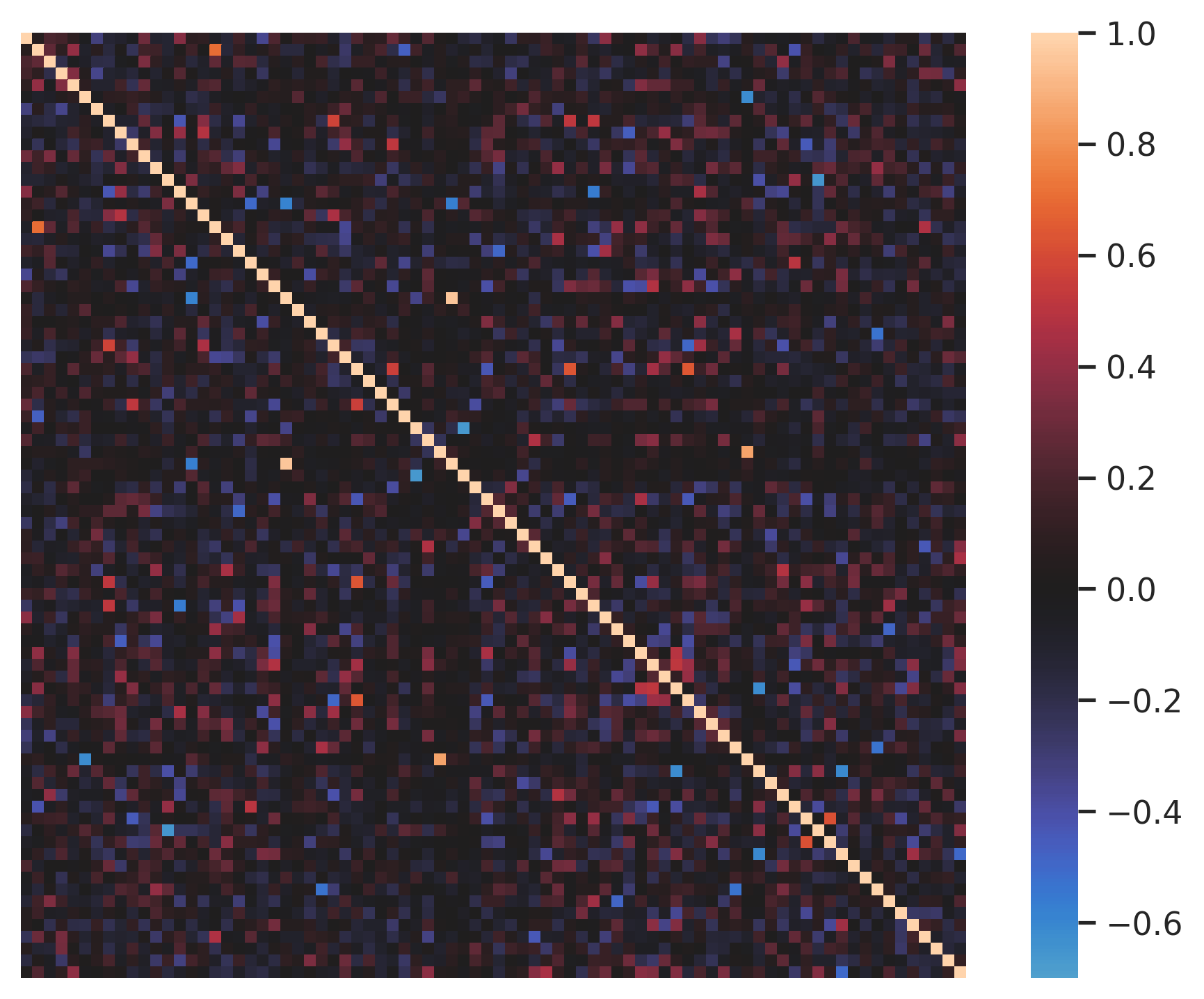}
			\includegraphics[height=0.23\linewidth]{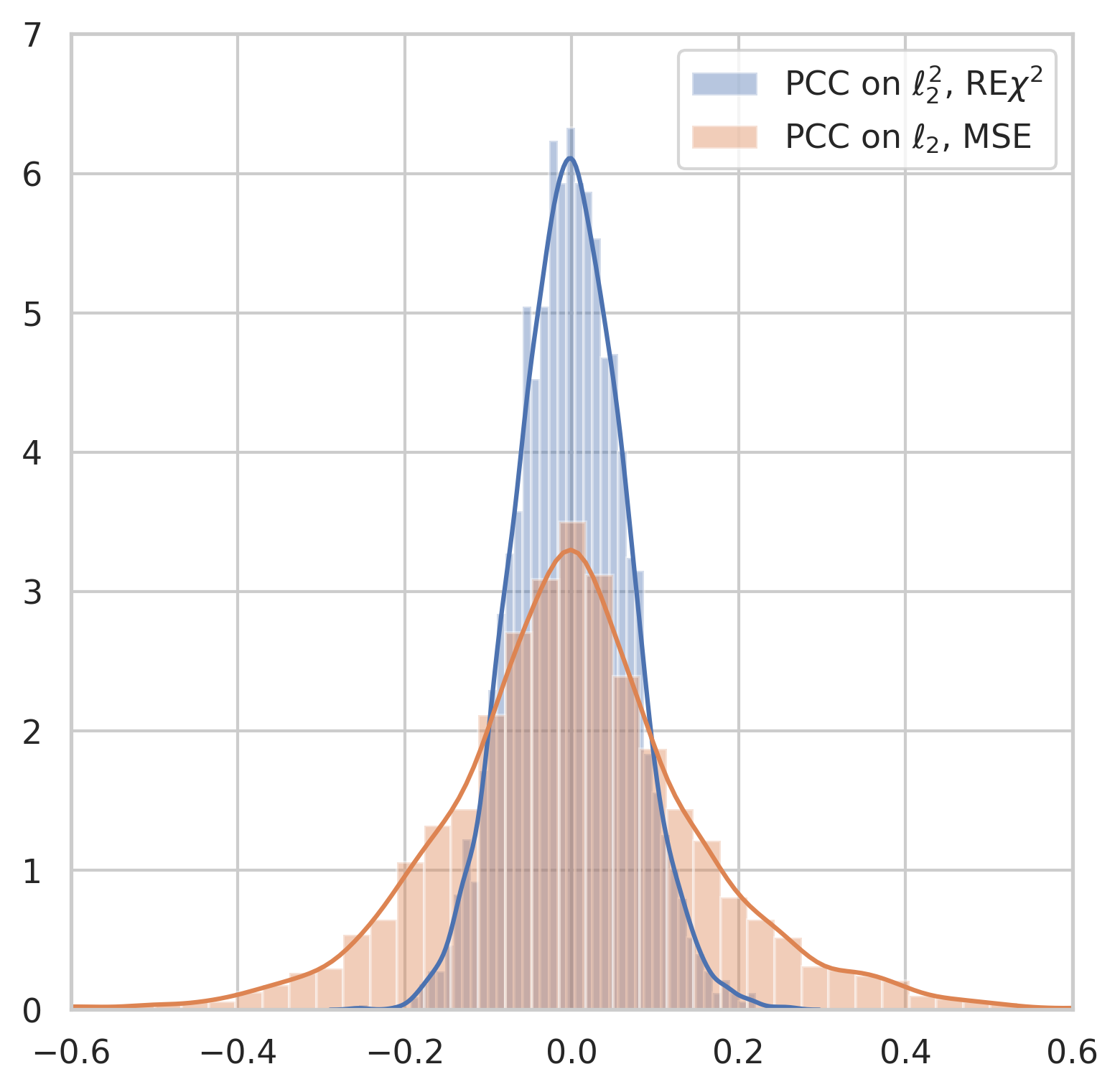}
		\caption{Heatmaps and distributions of the PCCs between different elements of embedding vectors. 
		The left subfigure is the heatmap for the proposed method (CNN-ED-5: $\ell_2^2$: RE$\chi^2$). 
		The middle subfigure is the heatmap for the comparative\slash ablation experiment (CNN-ED-5: $\ell_2$: MSE). 
		The right subfigure is for the histograms of the PCCs for the two models (diagonal $1$s are omitted).}
		\label{fig:pcc} 
		\end{center}
		\vskip -0.2in  
	\end{figure*}

The elements of the embeddings are produced by the same neural network. 
It is straightforward to suspect 
that the different embedding elements $u_i$ and $u_j$ are not independent in practice. 
However, as analyzed below, the proposed approach declines the dependence between the embedding elements 
and upholds the assumption that $u_i$ and $u_j$ are independent if $i\neq j$. 
Let $\bm{u}, \bm{v}$ be the rescaled embeddings of a pair of 
non-homologous sequences $\bm{s}, \bm{t}$, respectively. 
By the proposed squared Euclidean embedding and RE$\chi^2$ loss, 
the degree of freedom with the difference between the embedding vectors $\bm{u}-\bm{v}$ is encouraged to  
meet $d_L(\bm{s}, \bm{t})$, which is the ground truth Levenshtein distance between $\bm{s}, \bm{t}$. 
In view of this, the degrees of freedom with the non-homologous $\bm{u}-\bm{v}$ are approximately equal to 
the average Levenshtein distance between non-homologous pairs and equal to the embedding dimension, 
suggesting that the embedding elements tend to be independent of each other. 
To show the independence of the embedding elements experimentally, 
we plot the heatmap of the Pearson correlation coefficient (PCC) 
between $u_i$ and $u_j$ from the testing phase of the proposed method (CNN-ED-5: $\ell_2^2$: RE$\chi^2$) 
in the left subfigure of \cref{fig:pcc}\footnote{
	If $(X,Y)$ follows a bivariate normal distribution with covariance $0$, then $X,Y$ are independent.}. 
This heatmap shows that most PCCs ($i\neq j$) have absolute values less than $0.2$, 
indicating that $u_i$ and $u_j$ are weakly or not correlated. 
To illustrate the effectiveness of the proposed method, the heatmap from 
the comparative\slash ablation experiment 
(CNN-ED-5: $\ell_2$: MSE) is plotted in the middle subfigure of \cref{fig:pcc}. 
We also plot the histograms of the PCCs from the proposed method and the 
ablation study in the right subfigure of \cref{fig:pcc}. 
From these heatmaps and histograms, we find that the independence decreases 
in the absence of the squared Euclidean embedding and RE$\chi^2$ loss. 
This partly confirms our analysis that the proposed approach reduces the dependence between the different 
embedding elements.  

\end{document}